\documentclass[journal]{IEEEtran}
\usepackage[utf8]{inputenc}
\usepackage{cite}
\usepackage{hyperref}
\usepackage{multirow}
\usepackage{algorithmic}
\usepackage{orcidlink}
\usepackage{subfigure}
\usepackage{amsmath} 
\usepackage{amsfonts,amssymb}  
\usepackage{colortbl}
\usepackage{booktabs}
\usepackage{graphicx} 
\usepackage{titlesec} 
\titlespacing*{\section}{0pt}{0.3em}{0.3em}  
\titlespacing*{\subsection}{0pt}{0.2em}{0.2em}  
\usepackage{xcolor}
\graphicspath{ {./figures/} }
\begin{document}

\title{Clustering-Guided Spatial-Spectral Mamba for Hyperspectral Image Classification}

\author{Zack Dewis, Yimin Zhu, Zhengsen Xu, Mabel Heffring, Saeid Taleghanidoozdoozan, Quinn Ledingham, Lincoln Linlin Xu,~\IEEEmembership{Member,~IEEE}
\thanks{This work was supported by the Natural Sciences and Engineering Research Council of Canada (NSERC) under Grant RGPIN-2019-06744.}
\thanks{Lincoln Linlin Xu, Yimin Zhu, Zack Dewis, Zhengsen Xu, Mabel Heffring, Saeid Taleghanidoozdoozan, Quinn Ledingham, are all with the Department of Geomatics Engineering, University of Calgary, Canada (email: (lincoln.xu, yimin.zhu, zachary.dewis, zhengsen.xu, mabel.heffring1, quinn.ledingham, saeid.taleghanidoozd)@ucalgary.ca, ) (Corresponding author: Lincoln Linlin Xu).}
}

\markboth{Journal of \LaTeX\ Class Files,~Vol.~13, No.~9, September~2014}
{Shell \MakeLowercase{\textit{et al.}}: }

\maketitle
\begin{abstract}
Although Mamba models greatly improve Hyperspectral Image (HSI) classification, they have critical challenges in terms defining efficient and adaptive token sequences for improve performance. This paper therefore presents CSSMamba (Clustering-guided Spatial-Spectral Mamba) framework to better address the challenges, with the following contributions. First, to achieve efficient and adaptive token sequences for improved Mamba performance, we integrate the clustering mechanism into a spatial Mamba architecture, leading to a cluster-guided spatial Mamba module (CSpaMamba) that reduces the Mamba sequence length and improves Mamba feature learning capability. Second, to improve the learning of both spatial and spectral information, we integrate the CSpaMamba module with a spectral mamba module (SpeMamba), leading to a complete clustering-guided spatial-spectral Mamba framework.   Third, to further improve feature learning capability, we introduce an Attention-Driven Token Selection mechanism to optimize Mamba token sequencing. Last, to seamlessly integrate clustering into the Mamba model in a coherent manner, we design a Learnable Clustering Module that learns the cluster memberships in an adaptive manner. Experiments on the Pavia University, Indian Pines, and Liao-Ning 01 datasets demonstrate that CSSMamba achieves higher accuracy and better boundary preservation compared to state-of-the-art CNN, Transformer, and Mamba-based methods.
\end{abstract}


\IEEEpeerreviewmaketitle

\section{Introduction}
\vspace{-0.1cm}Hyperspectral image (HSI) classification, which transfers the raw HSI data into meaningful, pixel-level maps, is a critical task in various important applications, e.g., precision agriculture, environmental monitoring and urban planning. Nevertheless, it is a challenge task due to the high-dimensionality, the noise, and limited ground truth, which causes significant difficulties in learning the discriminative features in HSI \cite{Li2019}. Therefore, developing sophisticated feature learning frameworks that leverage advanced machine learning (ML) and deep learning (DL) models has become a critical research topic.
\begin{figure}[h]
    \centering
    \includegraphics[width=1\linewidth]{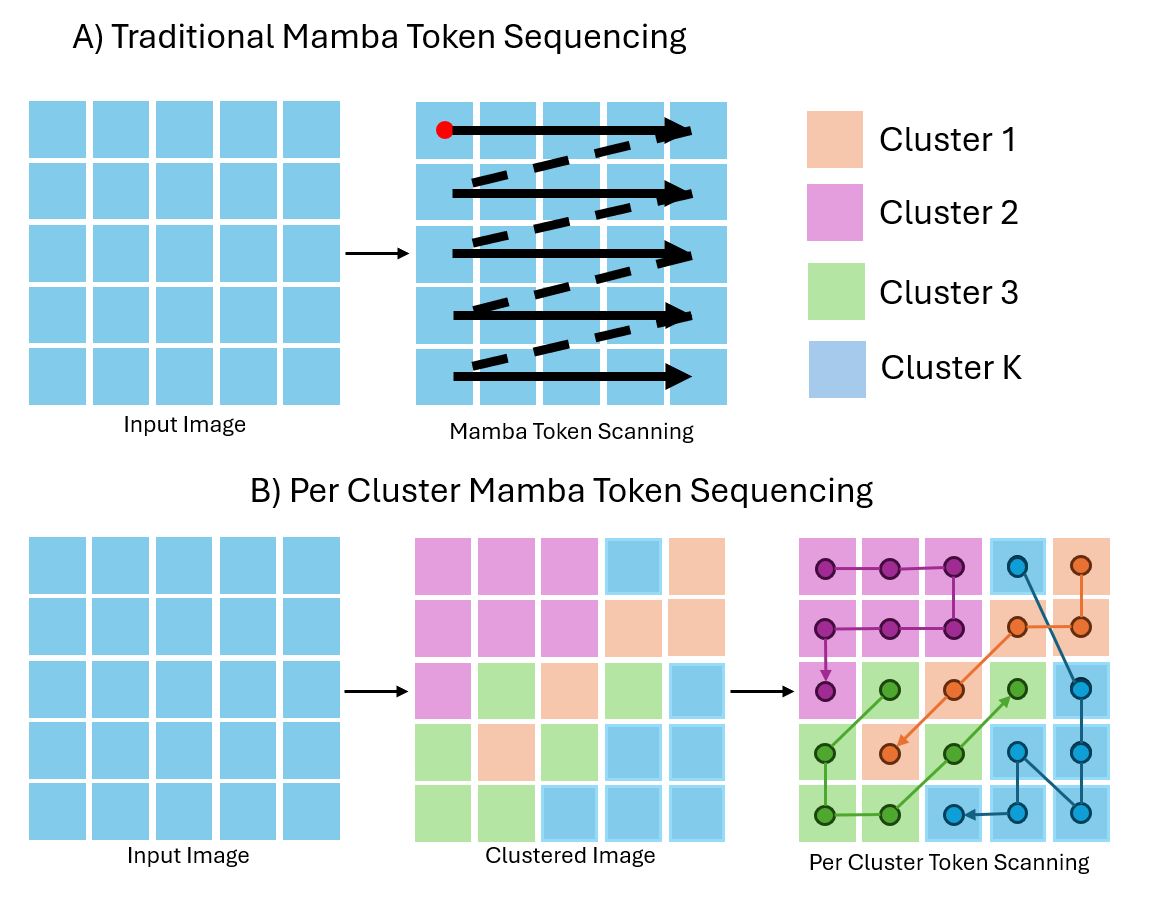}
    \caption{\textbf{Traditional Mamba token sequencing} (top) treats an entire image—despite its highly heterogeneous spatial patterns—as a single long sequence. This results in extremely long global token streams, which suffer from correlation decay and are inefficient at modeling fine-scale structures and subtle spatial variations.
      In contrast, our \textbf{Per-Cluster Mamba Token Sequencing approach} (bottom) decomposes a large, complex image into multiple smaller, locally homogeneous clusters to better capture subtle and weak patterns and edges, and constructs a dedicated token sequence for each cluster. This leads to significantly shorter sequences with more stationary feature distributions, making them easier to learn and better suited for capturing weak signals, fine details, and class boundaries. Furthermore, unlike conventional Mamba methods that scan tokens in a fixed, dense, and predefined order, our approach sequences pixels within each cluster using a \textbf{dynamic, learnable, sparse, and adaptive} strategy. This flexibility enables more effective modeling of local structures, improves detail preservation, and enhances the network’s ability to learn discriminative features from complex imagery.}

    \label{fig:perclusterscanning}
\end{figure}

Deep learning models have been widely used in HSI feature extraction to improve HSI classification. Convolutional Neural Networks (CNNs) can effectively capture local spatial-spectral features due to their hierarchical structure and shared weights for
convolutional operations. Nevertheless, CNNs are limited by their locality biases, preventing adequate modeling of long-range dependencies \cite{liu2022convnet}. To address the challenges posed by local context limitations, transformer architectures have emerged as a viable
alternative. They take advantage of self-attention mechanisms that allow models to focus on global features across entire images, which can significantly enhance the learning of large-scale spatial relationship within HSI classification tasks \cite{dosovitskiy2021image}. However, a key limitation of the transformer models is the large computational cost and the risk of overfitting due to the numerous parameters in the attention mechanism. 

Recently, the Mamba models have been widely used in improving the Transformer models in long-range large-scaling modeling application. Comparing with Transformer models, Mamba approaches model the tokens as a sequence with linear computational complexity ($O(N)$), greatly reducing the computational cost and thereby the overfitting risk \cite{11075710}. Nevertheless, how to construct Mamba token sequences is a critical issue that bottlenecks Mamba model performance. 

Traditional Mamba methods, as displayed in Figure \ref{fig:perclusterscanning}(A), treats an entire image—despite its highly heterogeneous spatial patterns—as a single long sequence. This results in extremely long global token streams, which suffer from correlation decay and are inefficient at modeling fine-scale structures and subtle spatial variations \cite{Stepchenko2017}. Although the patch-based Mamba, where each local patch is treated as a token, can reduce the number of tokens, they cause another serious problem by reducing the resolution and the valuable details in the classification map \cite{10604894}. As such, how to design a new Mamba token sequencing approach that can preserve the detail information while also improve Mamba feature learning is a critical research issue. 

This paper therefore presents a Clustering-guided Spatial-Spectral Mamba (CSSMamba) framework for improved HSI classification, with the following contributions. 


\begin{itemize}
\vspace{-0.1cm}
\item First, as shown in Figure \ref{fig:perclusterscanning}(B), to achieve efficient and adaptive token sequences for improved Mamba performance, instead of treating the complex and heterogeneous image as one sequence, our CSpaMamba decomposes a large, complex image into multiple smaller, locally homogeneous clusters and constructs a dedicated token sequence for each cluster. This leads to significantly shorter sequences with more stationary feature distributions, making them easier to learn and better suited for capturing weak signals, fine details, and class boundaries. 

\item Second, to improve the learning of both spatial and spectral information, we integrate the CSpaMamba module with a spectral mamba module (SpeMamba), leading to a complete clustering-guided spatial-spectral Mamba framework. 

\item Third, to further improve feature learning capability, we introduce a Dual-Attention Token Selection mechanism to sequence the tokens in a dynamic, learnable, sparse, and adaptive manner.

\item Last, to seamlessly integrate clustering into the Mamba model in a coherent manner, we design a Learnable Clustering Module that learns the cluster memberships in an adaptive manner. 

\end{itemize}


The remainder of the paper is organized as follows. Section II illustrates the details of the proposed CSSMamba approach. Section III presents the experimental design and results. Section IV concludes this study.

\section{Methodology}

Figure \ref{fig:overview} shows the architecture of the proposed CSSMamba model. We will illustrate its four building blocks below.

\subsection{Cluster-Guided Spatial Mamba Branch}
To effectively capture long-range spatial dependencies while preserving class boundaries, the spatial branch employs a \textit{Cluster-Guided Global-Local} architecture. Standard Mamba models process the flattened image sequence ($L=H \times W$) indiscriminately, often mixing distinct spectral signals (e.g., "water" vs. "urban") into a single state trajectory. Our module addresses this by decomposing the image into $K$ semantic groups and processing them through a parallelized dual-stage framework.

The input spatial tokens $\mathbf{X} \in \mathbb{R}^{B \times L \times D}$ (where $B$ is the batch size, $L$ is $H * W $, and $D$ are the features) first pass through a normalization layer and a linear projection to expand the feature dimension. The data flow then proceeds through three key stages:

\begin{enumerate}
    \item \textbf{Cluster-Based Partition:}
    Instead of processing the full grid, we partition the global tokens into a collection of $K$ distinct subsets based on the assignment indices from the Learnable Clustering Module. This results in a set of cluster-specific tensors $\mathcal{X}_{set} = \{ \mathbf{X_c}  | c = 1, 2, ..., K \}$, where $K$ is the total number of clusters. Here, any specific tensor $\mathbf{X}_c \in \mathbb{R}^{B \times N_c \times D}$ contains only the $N_c$ tokens assigned to the $c$-th semantic class. Note that $N_c$ varies dynamically per cluster (e.g., the number of "water" tokens differs from "urban" tokens). This also produces a cluster map $\mathbf{M} \in \mathbb{R}^{B \times L}$ which is later used in the dual attention module (Sec. C).

    \item \textbf{Cluster-Specific Guided Local Mamba Block:} 
    We employ $K$ independent Mamba blocks to process the partitioned set in parallel as seen in Figure \ref{fig:perclusterscanning}. Crucially, rather than processing the raw, unsorted tokens, we integrate our proposed token selection mechanism directly into the forward pass. For a specific cluster $c$, where $c \in \{1, \dots, K\}$, the processing is defined as:
    \begin{equation}
        \mathbf{Z}_c = \text{Mamba}^{(c)}\left( \text{DualAttn}(\mathbf{X}_c) \right)
    \end{equation}
    Here, $\text{DualAttn}(\cdot)$ applies the hybrid importance scoring (Sec. C) to sort and filter the input $\mathbf{X}_c$, returning a refined, information-dense sequence $\tilde{\mathbf{X}}_c \in \mathbb{R}^{B \times \bar{N} \times D}$ (where $\bar{N}$ is the number of tokens). By feeding this sorted sequence into $\text{Mamba}^{(c)}$, we ensure the state space model focuses its scanning trajectory on the most semantically salient nodes first, effectively reducing computational redundancy while enhancing feature extraction for the specific material type.

    \item \textbf{Global Mamba Block:} 
    The local processing step produces a set of disjoint feature tensors $\{ \mathbf{Z}_1, \dots, \mathbf{Z}_K \}$. To restore the global receptive field, these features are scattered back to their original spatial indices, reconstructing a dense feature map $\mathbf{Z}_{recon} \in \mathbb{R}^{B \times L \times D}$. This map is then combined with the original residual input and processed by a shared Global Mamba block:
    \begin{equation}
        \mathbf{Y} = \text{Mamba}_{global}\left( \text{}\left( \mathbf{X} + \mathbf{Z}_{recon} \right) \right)
    \end{equation}
    The final output $\mathbf{Y_{spa}} \in \mathbb{R}^{B \times L \times D}$ integrates the refined local features into a coherent whole, correcting potential boundary artifacts between clusters and ensuring spatial consistency across the entire scene.
\end{enumerate}

\subsection{Clustering-Guided Spatial Spectral Mamba Framework}
To comprehensively model the high-dimensional structure of hyperspectral data defined by the input $\mathbf{X} \in \mathbb{R}^{B \times C \times H \times W}$ (where $B$ is the batch size, $C$ is the number of spectral bands, and $H, W$ are the spatial dimensions), we propose a unified Spatial-Spectral Mamba framework as seen in Figure \ref{fig:overview}. The architecture prioritizes spectral fidelity through a dedicated \textbf{Spectral Branch}, which transforms the input into a flattened sequence of pixel-wise spectral tokens $\mathbf{T}_{spe} \in \mathbb{R}^{(B \cdot H \cdot W) \times N \times G}$. Here, $N$ represents the number of spectral tokens and $G$ denotes the number of channels per group, allowing the Mamba encoder to model long-range correlations across the entire spectral curve. Complementing this, a \textbf{Cluster-Guided Spatial Branch} efficiently captures geometric dependencies by routing spatial tokens based on semantic affinity. The framework concludes by fusing these complementary views, yielding a feature representation $\mathbf{Y} \in \mathbb{R}^{B \times C \times H \times W}$ that seamlessly integrates dense spectral signatures with global spatial structure.
\subsection{Dual Attention Module}

Standard State Space Models process sequences sequentially with uniform computational allocation, which is inefficient for HSI where background regions and redundant spectral information often dominate. Our framework employs an attention-driven mechanism to dynamically identify and sort the most informative tokens, processing them first, improving modeling accuracy.

Given the cluster-specific input tokens $\mathbf{X}_c \in \mathbb{R}^{B \times N_c \times D}$ (where $N_c$ is the number of tokens assigned to cluster $c$ and $D$ is the feature dimension), we compute a hybrid importance score $\mathbf{S}_{score} \in \mathbb{R}^{B \times \bar{N}}$ for each token. This score fuses two complementary information sources: dynamic self-attention that captures token-specific saliency, and a static cluster prior that ensures semantic coherence. The scoring mechanism is formalized as:
\begin{equation}
\scriptsize
    \mathbf{S}_{score} = \sigma(\alpha) \cdot \underbrace{\text{Softmax}\left(\frac{\mathbf{Q}\mathbf{K}^\top}{\sqrt{d}}\right) \cdot \mathbf{v}_{agg}}_{\text{Dynamic Attention}} + (1 - \sigma(\alpha)) \cdot \underbrace{\mathbf{P}_{cluster}(\mathbf{X}_c|\mathbf{M})}_{\text{Static Prior}}
\end{equation}
Here, $\mathbf{Q}, \mathbf{K} \in \mathbb{R}^{B \times \bar{N} \times d}$ are linear projections of the input tokens implementing scaled dot-product attention, $\mathbf{v}_{agg} \in \mathbb{R}^{B \times \bar{N} \times 1}$ is a learnable global context vector that aggregates cross-token information, and $\alpha$ is a learnable gating parameter. The cluster probability term $\mathbf{P}_{cluster}(\mathbf{X}_c|\mathbf{M}) \in \mathbb{R}^{B \times N_c}$ measures how representative each token is of its assigned cluster from $\mathbf{M}$, favoring tokens that are central to their semantic groups.

Based on the computed scores $\mathbf{S}_{score}$, we sort the most informative tokens to form a refined sequence $\tilde{\mathbf{X}}_c \in \mathbb{R}^{B \times \bar{N} \times D}$. This filtered sequence is fed into the subsequent mamba block.

\subsection{Learnable Clustering Module}
\begin{figure*}[htbp]
\centering
\includegraphics[scale=0.46]{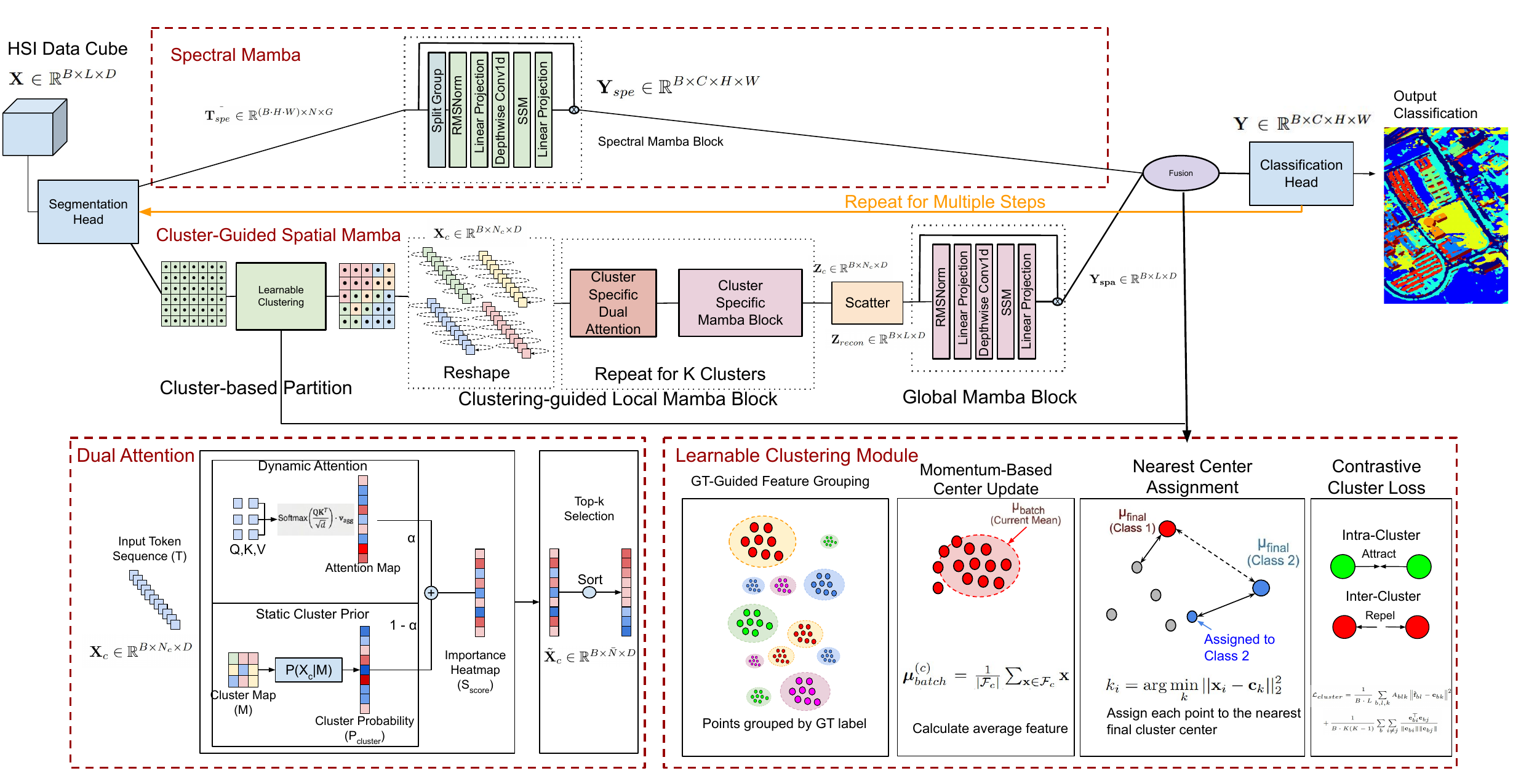}
\caption{The proposed \textbf{CSSMamba} is a Spatial-Spectral dual-branch architecture that features a  spectral branch and a cluster-guided spatial Mamba. The spatial branch leverages a learnable clustering mechanism, which reduces redundant computation by replacing dense grid scanning with dynamic, learnable, sparse and adaptable scanning. The Spatial-Spectral architecture is joined together to produce a feature map that jointly models dense photometric signatures and global semantic structure. A cluster-constrained dual loss guides the latent space to enforce intra-class compactness and inter-class separability, ensuring both spectral fidelity and spatial consistency. } 
\label{fig:overview}
\end{figure*}

To ensure the semantic decomposition utilized in Sec A remains robust and physically meaningful, we employ a Learnable Clustering Module that dynamically evolves the cluster centers to track the feature distribution. As illustrated in Figure \ref{fig:overview}, this module operates through a four-stage mechanism synchronized with the training iterations:

\begin{enumerate}
    \item \textbf{GT-Guided Feature Grouping:} 
    In each training step, we utilize the ground truth labels to group the feature representations $\mathbf{X}$ by class. For a specific semantic class $c$, we extract the set of valid feature vectors $\mathcal{F}_c = \{ \mathbf{x}_i \mid y_i = c \}$.

    \item \textbf{Momentum-Based Center Update:} 
    Unlike standard parameters updated via backpropagation, the cluster centers $\mathbf{C} \in \mathbb{R}^{K \times D}$ (where k is the number of tokens) are updated using feature statistics to ensure stability. We first compute the mean of the current batch features for each class: $\boldsymbol{\mu}_{batch}^{(c)} = \frac{1}{|\mathcal{F}_c|} \sum_{\mathbf{x} \in \mathcal{F}_c} \mathbf{x}$. The global cluster centers are then updated using an Exponential Moving Average (EMA):
    \begin{equation}
        \mathbf{c}_k^{(t)} \leftarrow \gamma \cdot \mathbf{c}_k^{(t-1)} + (1 - \gamma) \cdot \boldsymbol{\mu}_{batch}^{(c)}
    \end{equation}
    where $\gamma$ is a momentum coefficient. This ensures the centers smoothly track the evolving latent space without oscillating due to batch noise.

    \item \textbf{Nearest Center Assignment:} 
    Once the centers are updated, every pixel in the image (regardless of its ground truth) is assigned to its nearest semantic prototype to determine its cluster membership:
    \begin{equation}
        k_i = \arg\min_{k} || \mathbf{x}_i - \mathbf{c}_k ||_2^2
    \end{equation}
    This assignment generates the routing indices used by the Spatial Mamba branch to distribute tokens to the correct local processors.

    \item \textbf{Contrastive Cluster Loss:} 
    To structure the feature space, we compute a cluster loss that enforces intra-cluster compactness and inter-cluster orthogonality. Let $\hat{\mathbf{f}}_{bl} \in \mathbb{R}^D$ denote the $L_2$-normalized feature vector for the $l$-th token in batch $b$, and let $A_{blk}$ represent the soft assignment weight to cluster $k$. We first compute the cluster centers $\mathbf{c}_{bk}$ via the weighted average of assigned features. The loss function then simultaneously minimizes the weighted variance within each cluster and minimizes the pairwise cosine similarity between the centers of distinct clusters $\mathbf{c}_{bi}$ and $\mathbf{c}_{bj}$ (where $i \neq j$):
    
    \begin{equation}
    \tiny
        \mathcal{L}_{cluster} = \underbrace{\frac{1}{B \cdot L} \sum_{b,l,k} A_{blk} \left\| \hat{\mathbf{f}}_{bl} - \mathbf{c}_{bk} \right\|^2}_{\text{Intra-Cluster Variance}} + \underbrace{\frac{1}{B \cdot K(K-1)} \sum_{b} \sum_{i \neq j} \frac{\mathbf{c}_{bi}^\top \mathbf{c}_{bj}}{\| \mathbf{c}_{bi} \| \| \mathbf{c}_{bj} \|}}_{\text{Inter-Cluster Similarity}}
\end{equation}
\end{enumerate}

\begin{figure*}[ht]
\centering
\includegraphics[width=\textwidth]{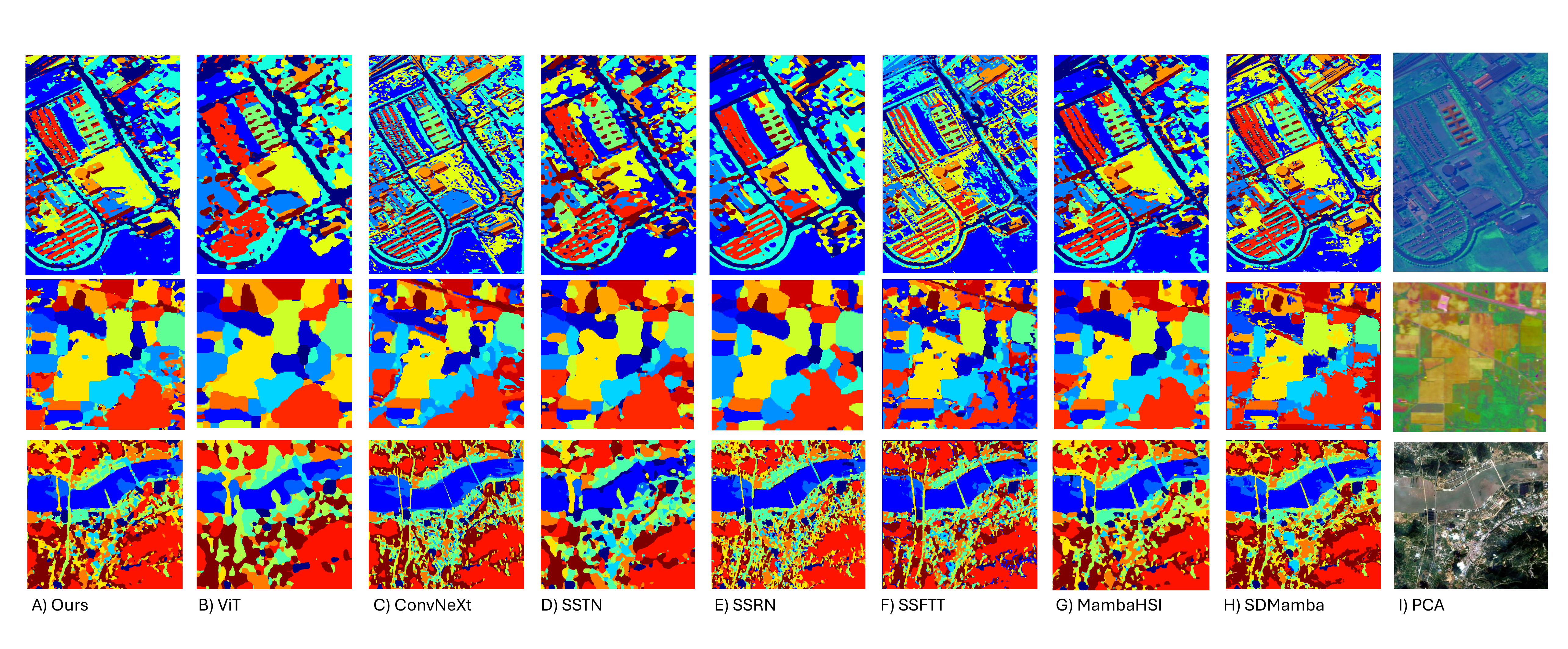}
\caption{The Indian Pines (middle), Pavia University (top) and, LN01 (bottom) classification map generated by different methods. (a) Ours (b) ViT (c) ConvNeXt (d) SSTN (e) SSRN (f) SSFTT (g) MammbaHSI (h) SDMamba (i) PCA images of the datasets}
\label{all_vis}
\end{figure*}


\section{Results and Analysis}
\subsection{Implementation schema}
\vspace{-0.1cm}We compare the proposed method with various state-of-the-art approaches, i.e., ViT \cite{dosovitskiy2021image}, ConvNeXt\cite{liu2022convnet}, SSTN \cite{zhongzlSSTN2022}, SSRN \cite{SSRN2018}, SSFTT\cite{SSFTT}, MambaHSI \cite{10604894} and, Sparse Deformable Mamba (SDMamba) \cite{11075710}, on three benchmark datasets, i.e., Pavia University (PU) (30 training pixels per class, 10 validation, the rest for testing), Indian Pines (IP) (30 training pixels per class, 10 validation, the rest for testing), LN01 dataset (30 training pixels per class, 10 validation, rest for testing). For classes with less than 40 samples, a 50/50 train test split was used, where a 3:1 ratio of train/val was kept for the training half. Overall accuracy (OA), averaged accuracy (AA), and the kappa coefficient for evaluating. For our method, the entire image is the input, a learning rate of $0.001$, $200$ epochs, $128$ for hidden dimensions. All training and testing was performed on a NVIDIA RTX A6000 Ada Generation with 48GB of VRAM using the PyTorch library.

\subsection{Results}
\definecolor{gold}{HTML}{D4AF37}
\definecolor{silver}{HTML}{C0C0C0}
\definecolor{bronze}{HTML}{CD7F32}

\newcommand{\first}[1]{\textcolor{gold}{\textbf{#1}}}
\newcommand{\second}[1]{\textcolor{silver}{\textbf{#1}}}
\newcommand{\third}[1]{\textcolor{bronze}{\textbf{#1}}}

\begin{table}[htbp]
\centering
\caption{Classification results for the Pavia University (PaviaU) dataset.}
\resizebox{\linewidth}{!}{%
\scriptsize
\setlength{\tabcolsep}{2.5pt}
\begin{tabular}{l@{\hspace{3pt}}l@{\hspace{3pt}}c@{\hspace{3pt}}c@{\hspace{3pt}}c@{\hspace{3pt}}c@{\hspace{3pt}}c@{\hspace{3pt}}c@{\hspace{3pt}}c@{\hspace{3pt}}c}
\toprule
\textbf{No.} & \textbf{Class Name} & \textbf{Ours} & \textbf{ViT} & \textbf{ConvNeXt} & \textbf{SSTN} & \textbf{SSRN} & \textbf{SSFTT} & \textbf{MambaHSI} & \textbf{SDMamba} \\
\midrule
C1 & Asphalt & \first{97.16} & 81.79 & 86.04 & 82.60 & \third{91.06} & 90.53 & \second{92.94} & 86.24 \\
C2 & Meadows & \second{97.99} & \first{98.22} & 89.56 & 84.82 & 80.87 & \third{95.56} & 93.57 & 78.44 \\
C3 & Gravel & 90.43 & \second{94.94} & 87.27 & 79.55 & 86.11 & \third{94.27} & \first{96.84} & 83.68 \\
C4 & Trees & 97.45 & 82.70 & \first{99.63} & 79.70 & \second{99.07} & 94.31 & 90.57 & \third{99.07} \\
C5 & Painted metal sheets & \first{100.00} & 97.47 & \first{100.00} & \second{99.92} & \first{100.00} & \third{99.46} & \first{100.00} & \first{100.00} \\
C6 & Bare Soil & \second{99.01} & \second{99.01} & 74.10 & 86.99 & \third{97.45} & 37.18 & \first{99.27} & 96.85 \\
C7 & Bitumen & \third{99.76} & 98.75 & 99.37 & 79.92 & \first{100.00} & \second{99.77} & 97.67 & 96.20 \\
C8 & Self-Blocking Bricks & \first{95.88} & 92.94 & 71.69 & 91.95 & 92.17 & \second{95.39} & \third{95.33} & 91.63 \\
C9 & Shadows & \third{99.88} & 92.50 & \first{100.00} & 97.24 & \first{100.00} & \second{99.89} & 99.69 & \first{100.00} \\
\midrule
OA & - & \first{97.55} & \third{93.91} & 87.11 & 85.31 & 88.51 & 88.81 & \second{94.69} & 86.34 \\
AA & - & \first{97.51} & 93.15 & 89.74 & 86.91 & \third{94.08} & 89.60 & \second{96.21} & 92.46 \\
Kappa & - & \first{98.75} & \second{95.00} & \third{91.25} & 87.50 & 85.31 & 84.00 & \first{98.75} & 82.57 \\
\bottomrule
\end{tabular}%
}
\label{PU}
\end{table}

\begin{table}[htbp]
\centering
\caption{Classification results for the Indian Pines (IP) dataset.}
\resizebox{\linewidth}{!}{%
\scriptsize
\setlength{\tabcolsep}{2.5pt}
\begin{tabular}{l@{\hspace{3pt}}l@{\hspace{3pt}}c@{\hspace{3pt}}c@{\hspace{3pt}}c@{\hspace{3pt}}c@{\hspace{3pt}}c@{\hspace{3pt}}c@{\hspace{3pt}}c@{\hspace{3pt}}c}
\toprule
\textbf{No.} & \textbf{Class Name} & \textbf{Ours} & \textbf{ViT} & \textbf{ConvNeXt} & \textbf{SSTN} & \textbf{SSRN} & \textbf{SSFTT} & \textbf{MambaHSI} & \textbf{SDMamba} \\
\midrule
C1 & Alfalfa & \second{83.33} & \first{100.00} & \first{100.00} & \second{83.33} & \first{100.00} & \first{100.00} & \first{100.00} & \first{100.00} \\
C2 & Corn-notill & \third{89.69} & 83.71 & 74.49 & 81.12 & \first{92.22} & 79.11 & \second{92.14} & 59.87 \\
C3 & Corn-mintill & \first{94.93} & 89.36 & 80.37 & 86.96 & \second{94.18} & 85.06 & \third{91.39} & 72.66 \\
C4 & Corn & \third{97.96} & \first{100.00} & \first{100.00} & 96.45 & \first{100.00} & 94.92 & \second{98.40} & 97.46 \\
C5 & Grass-pasture & \second{94.13} & \first{95.93} & 92.32 & 88.49 & 92.78 & \third{93.91} & 93.90 & 93.45 \\
C6 & Grass-trees & \second{99.56} & 95.50 & 93.33 & 93.62 & 97.97 & \first{99.57} & \third{98.98} & 98.41 \\
C7 & Grass-pasture-mowed & \second{92.85} & \first{100.00} & \first{100.00} & \first{100.00} & \first{100.00} & \first{100.00} & \second{92.85} & \first{100.00} \\
C8 & Hay-windrowed & \second{99.77} & \first{100.00} & \first{100.00} & \first{100.00} & \first{100.00} & 98.40 & \second{99.77} & \third{99.32} \\
C9 & Oats & \first{100.00} & \first{100.00} & \first{100.00} & \first{100.00} & \first{100.00} & \first{100.00} & \first{100.00} & \first{100.00} \\
C10 & Soybean-notill & 90.34 & \second{95.06} & 78.96 & \third{94.96} & \first{96.24} & 84.87 & 24.80 & 85.62 \\
C11 & Soybean-mintill & \first{95.90} & \third{90.39} & 74.78 & \second{92.05} & 90.19 & 82.53 & 85.96 & 73.87 \\
C12 & Soybean-clean & \first{95.84} & 82.09 & 79.92 & 86.44 & \third{89.69} & 67.09 & \second{92.04} & 77.22 \\
C13 & Wheat & \first{100.00} & \first{100.00} & \first{100.00} & \first{100.00} & \first{100.00} & \first{100.00} & \first{100.00} & \first{100.00} \\
C14 & Woods & \second{98.85} & \first{99.67} & 94.20 & 97.55 & 97.47 & 85.70 & \third{98.44} & 94.61 \\
C15 & Buildings-Grass-Trees & \first{100.00} & 99.13 & 97.10 & 95.38 & \third{99.42} & 83.82 & \second{99.71} & \second{99.71} \\
C16 & Stone-Steel-Towers & \first{100.00} & \first{100.00} & \first{100.00} & \first{100.00} & \first{100.00} & \first{100.00} & \second{98.11} & \first{100.00} \\
\midrule
OA & - & \first{95.39} & 92.30 & 83.54 & 91.43 & \second{94.19} & 84.50 & \third{92.86} & 81.56 \\
AA & - & \third{95.82} & 95.68 & 90.39 & 93.52 & \first{96.88} & 89.27 & \second{95.89} & 90.76 \\
Kappa & - & \second{98.04} & \first{98.56} & 93.84 & \third{97.85} & 93.36 & 82.29 & 96.74 & 78.96 \\
\bottomrule
\end{tabular}%
}
\label{IP}
\end{table}

\begin{table}[htbp]
\centering
\caption{Classification results for the LN01 dataset.}
\resizebox{\linewidth}{!}{%
\scriptsize
\setlength{\tabcolsep}{2.5pt}
\begin{tabular}{l@{\hspace{3pt}}l@{\hspace{3pt}}c@{\hspace{3pt}}c@{\hspace{3pt}}c@{\hspace{3pt}}c@{\hspace{3pt}}c@{\hspace{3pt}}c@{\hspace{3pt}}c@{\hspace{3pt}}c}
\toprule
\textbf{No.} & \textbf{Class Name} & \textbf{Ours} & \textbf{ViT} & \textbf{ConvNeXt} & \textbf{SSTN} & \textbf{SSRN} & \textbf{SSFTT} & \textbf{MambaHSI} & \textbf{SDMamba} \\
\midrule
C1 & Reservoir & \second{98.56} & 90.64 & \third{98.20} & 89.03 & \first{99.64} & 97.12 & 96.58 & 92.63 \\
C2 & Seawater & 89.07 & 73.92 & \first{91.10} & 82.67 & \second{90.38} & \third{89.26} & 84.37 & 88.52 \\
C3 & Sandy soil & \third{90.27} & 76.38 & \first{91.34} & 41.97 & 83.38 & 76.35 & 89.96 & \second{91.15} \\
C4 & Broken bridge & \first{100.00} & \first{100.00} & \first{100.00} & \third{31.82} & \first{100.00} & \first{100.00} & \first{100.00} & \second{95.45} \\
C5 & Barren grass & 71.84 & 68.81 & \third{85.74} & 78.97 & 83.79 & \second{86.65} & 84.40 & \first{93.51} \\
C6 & Highway & \first{88.55} & 63.68 & 58.54 & 63.98 & \second{73.30} & \second{73.30} & 60.53 & \third{71.81} \\
C7 & Railway & \third{90.76} & 86.15 & \second{93.84} & 80.00 & \first{95.38} & \first{95.38} & 78.46 & 87.69 \\
C8 & Bare soil & \second{96.76} & 94.15 & 91.07 & 87.08 & \first{97.69} & 87.69 & \third{94.30} & 88.77 \\
C9 & Mountain vegetation & \first{98.16} & 89.87 & 91.55 & 92.88 & \second{97.22} & 93.63 & 94.21 & \third{94.90} \\
C10 & Arable land & \second{89.41} & 71.44 & 82.61 & 69.50 & 59.54 & 80.18 & \third{88.73} & \first{92.23} \\
\midrule
OA & - & \first{91.77} & 80.53 & 89.82 & 82.22 & \third{89.89} & 89.14 & 88.53 & \second{90.68} \\
AA & - & \first{91.34} & 82.50 & \third{88.55} & 69.09 & 88.03 & 87.96 & 87.15 & \second{88.67} \\
Kappa & - & \first{94.44} & 74.44 & \second{87.77} & 63.33 & 84.44 & 84.96 & 83.33 & \third{87.19} \\
\bottomrule
\end{tabular}%
}
\label{LN}
\end{table}

\begin{table}[htbp]
\centering
\caption{Ablation study results: Number of Clusters, Loss Function, and Attention Mechanism.}
\begin{tabular}{lccc}
\toprule
\textbf{Condition} & \textbf{OA (\%)} & \textbf{AA (\%)} & \textbf{Kappa } \\
\midrule
\multicolumn{4}{l}{\textit{Number of Clusters per Class}} \\
1 & 96.01 & 96.71 & \textbf{98.75} \\
2 & 96.34 & 97.40 & \textbf{98.75} \\
3 & \textbf{97.55} & \textbf{97.51} & \textbf{98.75} \\
4 & 96.53 & 96.69 & 95.00 \\
5 & 96.40 & 97.33 & 96.25 \\
\midrule
\multicolumn{4}{l}{\textit{Loss Function}} \\
Cluster Loss, Center Updates + CE & \textbf{97.55} & \textbf{97.51} & \textbf{98.75} \\
CE Only & 96.71 & 96.73 & 97.50 \\
\midrule
\multicolumn{4}{l}{\textit{Attention Mechanism}} \\
With Attention & \textbf{97.55} & \textbf{97.51} & \textbf{98.75} \\
Without Attention & 97.15 & 97.18 & \textbf{98.75} \\
\bottomrule
\end{tabular}
\label{ablation}
\end{table}
\vspace{-0.1cm}Figure \ref{all_vis} shows the classification maps achieved by different methods on the IP, PU, and LN01 datasets. As we can see, the proposed approach achieves a map that is not only the most consistent with the classification map referencing to the RGB image, but also better at delineating the boundaries and small classes. SSRN, SSTN and, ViT tend to have smooth boundaries, leading to a lack of detail. SDMamba tends to have fragmented classification results, which may be due to the insufficient fusion of multi-scale spatial features due to its sparsity.

Table \ref{PU}, \ref{IP}, and \ref{LN} show the numerical results achieved by different methods on the three datasets. Our approach outperforms the other methods on all metrics for the LN and PU datasets. While for IP we have the highest OA, second Kappa and third AA. In particular, our approach achieves much better results on AA, indicating that the proposed approach outperforms the other approaches in terms of preserving and classifying the small classes. 

Furthermore, Table \ref{ablation} shows the results of the ablation studies. It can be seen that 3 clusters per class is the optimal number, achieving the highest accuracy. The Cluster Loss introduced in this paper also leads to a higher accuracy than just using a standard cross entropy loss. Using the dual attention module also leads to an accuracy boost to order the tokens instead of using a simple linear layer for ordering.

\section{Conclusion}
\vspace{-0.1cm}
In this paper, we presented the CSSMamba  framework to address critical challenges in defining efficient and adaptive token sequences for HSI classification. First, we developed a Cluster-Guided Spatial Mamba (CSpaMamba) module by integrating a clustering mechanism into the spatial Mamba architecture, which effectively reduces Mamba sequence length and enhances feature learning capability. Second, we constructed a complete clustering-guided spatial-spectral framework by integrating CSpaMamba with a Spectral Mamba (SpeMamba) module to simultaneously improve the learning of spatial and spectral information. Third, we introduced an Dual Attention Module to optimize Mamba token sequencing and feature extraction. Last, a Learnable Clustering Module that learns cluster memberships in an adaptive manner. Extensive experiments on multiple benchmark datasets demonstrate that the proposed method achieves higher accuracy and superior boundary preservation compared to state-of-the-art CNN, Transformer, and Mamba-based methods.
\bibliographystyle{IEEEtran}
\bibliography{IEEEabrv,references}

@article{Li2019,
  author    = {Li, S. and Song, W. and Fang, L. and Chen, Y. and Ghamisi, P. and Benediktsson, J.},
  title     = {Deep learning for hyperspectral image classification: an overview},
  journal   = {IEEE Transactions on Geoscience and Remote Sensing},
  year      = {2019},
  volume    = {57},
  number    = {9},
  pages     = {6690--6709},
  doi       = {10.1109/TGRS.2019.2907932}
}

@inproceedings{Stepchenko2017,
  author = {Stepchenko, A.},
  title = {Land cover classification based on MODIS imagery data using artificial neural networks},
  booktitle = {Environment Technology Resources Proceedings of the International Scientific and Practical Conference},
  year = {2017},
  volume = {2},
  pages = {159},
  doi = {10.17770/etr2017vol2.2545}
}

@ARTICLE{11075710,
  author={Xu, Lincoln Linlin and Zhu, Yimin and Dewis, Zack and Xu, Zhengsen and Alkayid, Motasem and Heffring, Mabel and Taleghanidoozdoozan, Saeid},
  journal={IEEE Geoscience and Remote Sensing Letters}, 
  title={Sparse Deformable Mamba for Hyperspectral Image Classification}, 
  year={2025},
  volume={22},
  number={},
  pages={1-5},
  doi={10.1109/LGRS.2025.3587256}}

@ARTICLE{10604894,
  author={Li, Yapeng and Luo, Yong and Zhang, Lefei and Wang, Zengmao and Du, Bo},
  journal={IEEE Transactions on Geoscience and Remote Sensing}, 
  title={MambaHSI: Spatial–Spectral Mamba for Hyperspectral Image Classification}, 
  year={2024},
  volume={62},
  number={},
  pages={1-16},
  doi={10.1109/TGRS.2024.3430985}}

@ARTICLE{SSFTT,
  author={Sun, Le and Zhao, Guangrui and Zheng, Yuhui and Wu, Zebin},
  journal={IEEE Transactions on Geoscience and Remote Sensing}, 
  title={Spectral–Spatial Feature Tokenization Transformer for Hyperspectral Image Classification}, 
  year={2022},
  volume={60},
  number={},
  pages={1-14},
  doi={10.1109/TGRS.2022.3144158}}

@ARTICLE{SSRN2018,
 
  author={Zhong, Zilong and Li, Jonathan and Luo, Zhiming and Chapman, Michael},
 
  journal={IEEE Transactions on Geoscience and Remote Sensing},
 
  title={Spectral–Spatial Residual Network for Hyperspectral Image Classification: A 3-D Deep Learning Framework},
 
  year={2018},
 
  volume={56},
 
  number={2},
 
  pages={847-858},
}

@ARTICLE{zhongzlSSTN2022,
  author={Zhong, Zilong and Li, Ying and Ma, Lingfei and Li, Jonathan and Zheng, Wei-Shi},
  journal={IEEE Transactions on Geoscience and Remote Sensing}, 
  title={Spectral–Spatial Transformer Network for Hyperspectral Image Classification: A Factorized Architecture Search Framework}, 
  year={2022},
  volume={60},
  number={},
  pages={1-15}}

@inproceedings{dosovitskiy2021image,
  title={An Image is Worth 16x16 Words: Transformers for Image Recognition at Scale},
  author={Dosovitskiy, Alexey and Beyer, Lucas and Kolesnikov, Alexander and Weissenborn, Dirk and Zhai, Xiaohua and Unterthiner, Thomas and Dehghani, Mostafa and Minderer, Matthias and Heigold, Georg and Gelly, Sylvain and Uszkoreit, Jakob and Houlsby, Neil},
  booktitle={International Conference on Learning Representations},
  year={2021},
  url={https://openreview.net/forum?id=YicbFdNTTy}
}

@inproceedings{liu2022convnet,
  title={A ConvNet for the 2020s},
  author={Liu, Zhuang and Mao, Hanzi and Wu, Chao-Yuan and Feichtenhofer, Christoph and Darrell, Trevor and Xie, Saining},
  booktitle={Proceedings of the IEEE/CVF Conference on Computer Vision and Pattern Recognition (CVPR)},
  pages={11976--11986},
  year={2022}
}

\end{document}